\documentclass{IEEEtran}
\usepackage{cite}
\usepackage{amsmath,amssymb,amsfonts}
\usepackage{algorithm,algorithmic}
\usepackage{graphicx}
\usepackage{textcomp}
\usepackage{xcolor}
\def\BibTeX{{\rm B\kern-.05em{\sc i\kern-.025em b}\kern-.08em
    T\kern-.1667em\lower.7ex\hbox{E}\kern-.125emX}}

\begin{document}

\title{Localized Compression: Applying Convolutional Neural Networks to Compressed Images}
\author{Christopher A. George and Bradley M. West
\thanks{C. A. George and B. West are with Boston Fusion Corp., 70 Westview Street, Suite 100, Lexington, MA, 02421 (e-mail: alex.george@bostonfusion.com; bradley.west@bostonfusion.com)}
}

\maketitle

\begin{abstract}
We address the challenge of applying existing convolutional neural network (CNN) architectures to compressed images. Existing CNN architectures represent images as a matrix of pixel intensities with a specified dimension; this desired dimension is achieved by downgrading or cropping. Downgrading and cropping are attractive in that the result is also an image; however, an algorithm producing an alternative ``compressed'' representation could yield better classification performance. This compression algorithm need not be reversible, but must be compatible with the CNN's operations. This problem is thus the counterpart of the well-studied problem of applying compressed CNNs to uncompressed images, which has attracted great interest as CNNs are deployed to size-, weight-, and power- (SWaP)-limited devices. In this brief, we introduce \textit{Localized Compression}, a generalization of downgrading in which the original image is divided into blocks and each block is compressed to a smaller size using either sampling- or random-matrix-based techniques. By aligning the size of the compressed blocks with the size of the CNN's convolutional region, localized compression can be made compatible with any CNN architecture. Our experimental results show that Localized Compression results in classification accuracy approximately 1-2\% higher than is achieved by downgrading to the equivalent resolution.
\end{abstract}


\pagenumbering{gobble}

\section{Introduction}
\label{sec:introduction}
Convolutional Neural Networks (CNNs) \cite{DLbook} are the state-of-the-art technique for image classification, routinely achieving better-than-human performance. New CNN architectures and applications continue to emerge at a prodigious rate. More recently, substantial interest has arisen in compressing neural networks, including CNNs, to use fewer parameters and to require less memory so as to enable running on devices with limited size, weight, and power (SWaP). Note, ``compression'' in this context refers to reducing these computational and memory requirements while minimizing the effect on classification accuracy; this does not necessarily require that the compression can be reversed.

Compressing the network, however, addresses only one side of the coin: what about compressing the \textit{images} to which the CNN is applied? Though images are often stored in compressed form, CNN architectures currently uncompress all images prior to classifying them. Being able to compress the images also presents an additional advantage: given a dataset of large images and a network that expects small images, such a compression algorithm may preserve more information than extant techniques such as downgrading (DG) or cropping. Thus, this brief presents a compression algorithm that reduces the images' size on disk and does not require (or even allow) the images to be uncompressed prior to being classified by the CNN. 

\begin{figure*}
  \includegraphics[width=5.95in]{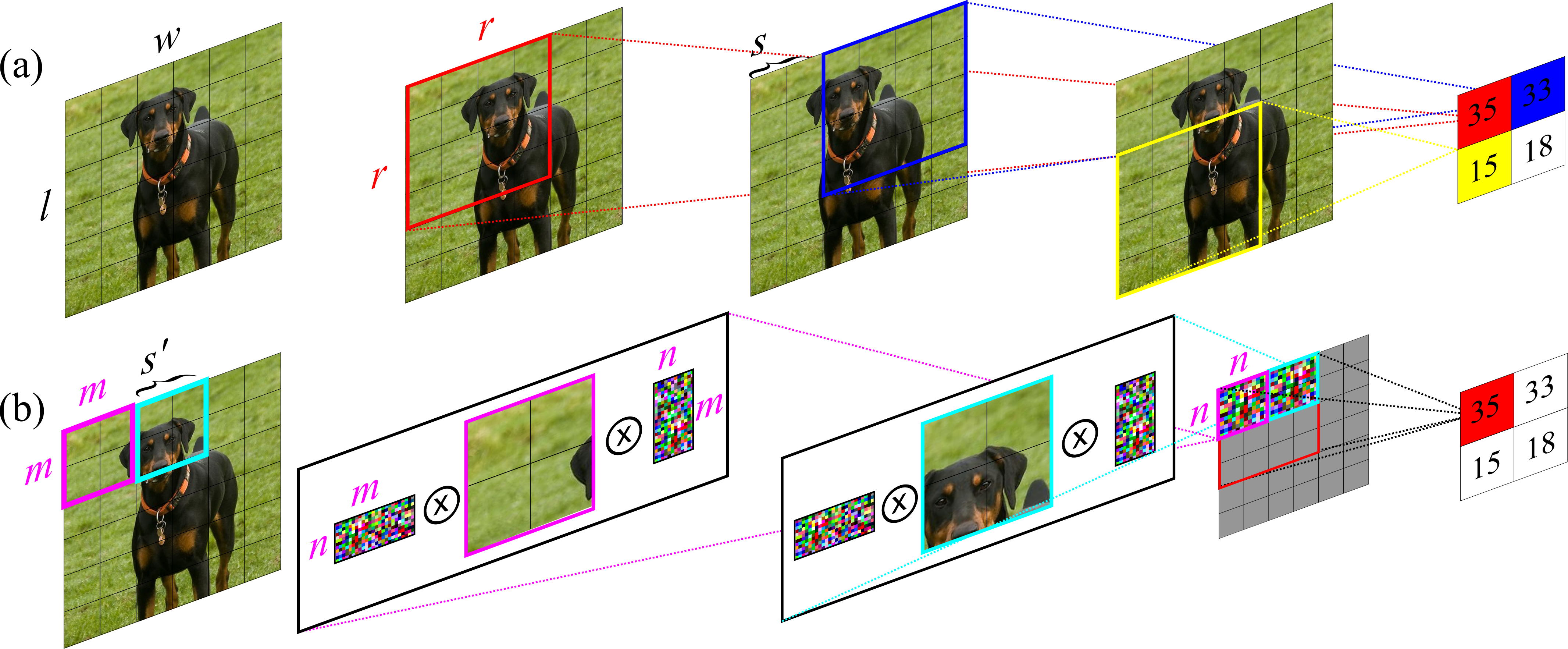}
  \centering
  \caption{Illustration of Localized Compression performed with Matrix Sketching}
  \label{fig:dimensions}
\end{figure*}

To understand why extant compression algorithms are inadequate, we we must consider how the CNN ingests the original $l \times w \times c$ image. The first layer of a CNN begins by ingesting a small $r \times r \times c$ ``convolutional region'' from the top-left of the image (the value of $r$ is set by the CNN architecture). After processing this area, the convolutional region ``strides'' (is translated) $s$ pixels to the right and the process repeats; in this way, the convolutional region ``convolves'' left-to-right, top-to-bottom across the matrix of pixel intensities (see Figure \ref{fig:dimensions}a). Thus, any effective compression scheme must preserve the localization, such that nearby pixels generally correspond to semantically coherent information. It is this requirement that existing techniques, such as JPEG compression, fail to meet. 

In response, we propose ``Localized Compression'' (LC). Rather than compressing the image as a whole, we divide the original image into $m \times m$ blocks and compress each block to $n \times n$ (with $n < m$). This reduces the number of pixels in the compressed image by a factor of $n^2/m^2$. While there are no restrictions on $m$, we require $n$ to be chosen such that $s$ is divisible by $n$; this ensures that each convolutional region receives the compresed pixels in the same relative order. This is illustrated in Figure \ref{fig:dimensions}b.

In principal, we could use standard compression techniques like JPEG compression or Principal Component Analysis (PCA) to compress these $m \times m$ blocks to $n \times n$ blocks. In practice, however, most modern CNN architectures have very small values of $s$, such as 4 (AlexNet), or even 2 (DenseNet). We must therefore compress already small blocks (e.g., $m=7$) into much smaller blocks (e.g., $n=s=2$). This rules out many well-established (reversible) compression techniques, including PCA and JPEG compression. 

Instead, we consider two solutions that are compatible with such small sizes: random matrix multiplication (RMM) and percentile-based sampling. RMM entails multiplying the original $m \times m$ matrix by random matrices of the appropriate dimensions; this has proven effective on related problems \cite{Thak}, \cite{sketching}, \cite{RandProjPriorWork}. Percentile-based sampling techniques entail retaining the maximum, minimum, and other values from the original matrix. 

We evaluate LC and compare these different options using to two standard datasets, ImageNet \cite{ImageNet} and the German Traffic Sign Recognition Benchmark \cite{GTSRB}, and two different CNN architectures, AlexNet \cite{AlexNet} and DenseNet \cite{DenseNet}. Both LC and DG produce images of the same size and therefore offer the same reduction in storage and processing requirements; therefore, we will compare them in terms of classification accuracy. Our results show that LC with percentile-based sampling is approximately 2\% more accurate than DG when $m \gg n$.

The remainder of this brief is organized as follows. Section \ref{sec:Related Work} describes related work. Section \ref{sec:Theory} provides more detail on random-matrix-based, sampling-based, and other techniques to represent an $m \times m$ matrix with an $n \times n$ matrix, while Section \ref{sec:Method} formally defines using these techniques for LC. Section \ref{sec:Experiments} shows numerical results from applying LC to standard datasets with existing network architectures. Section \ref{sec:Digression} contains a brief digression in which we apply random matrices to the related problem of compressing fully-connected layers. Finally, we draw conclusions in Section \ref{sec:Conclusions}.

\section{Related Work}
\label{sec:Related Work}

To our knowledge, there is no past work that addresses applying CNNs to compressed images (i.e., without immediately uncompressing each image). There is, however, much related work about compressing the CNN itself, and about compressing the inputs to other types of classifiers.

With respect to compressing the CNN itself, research has focused on four key areas: (1) reducing the number of network parameters by pruning and sharing, (2) using low-rank factorization to compress the network weights, (3) representing convolutional filters as transformations of a small number of base filters, and (4) transferring the essential knowledge from a deep network to a shallower network (knowledge distillation) \cite{deepcompression}, \cite{survey}. Of these, using low-rank factorization to compress the network weights is most similar to our paradigm. In particular, Denton \textit{et al.} \cite{Denton} showed that performing tensor decompositions (based on the singular value decomposition) on trained convolutional layers can significantly accelerate CNNs with minimal loss in classification accuracy. While we also consider extensions of the singular value decomposition (Section \ref{sec:Theory}), our work is different in that we compress the inputs to the CNN (images) rather than the CNN itself.

Compressing inputs to simple classifiers (single-layer perceptrons) has also been thoroughly studied. In particular, Wimalajeewa and Varshney \cite{Thak} recently considered sparse random matrices to compress the input to simple classifiers (nearest neighbor classifiers, random forests, and support vector machines), and W{\'o}jcik \textit{et al.}\ \cite{RandProjPriorWork} studied compressing high-dimensional vector input (e.g., telemetry data) to deep, fully-connected neural networks. While we also consider random matrices, we apply them to CNNs rather than simple classifiers or fully-connected neural networks.

Section \ref{sec:Digression} discusses using random matrices to compress a CNN's fully-connected layers. Here, there is substantial related work: in particular, Cheng\ \textit{et al.}\ \cite{circulant} has shown that circulant projection matrices (a subset of random projection matrices) efficiently compress fully-connected layers in CNNs, and W\'ojcik \textit{et al.} \cite{RandProjPriorWork} considers random projection matrices for the same purpose in fully-connected (non-convolutional) neural networks. Our work on the fully-connected layers fills in the gap, using random, non-circulant matrices to compress fully-connected layers in CNNs.

\section{Theory}
\label{sec:Theory}

We begin by considering how to compress two-dimensional $m \times m$ blocks to $n \times n$ with $n < m$ (in this work, we treat each channel separately). We refer to the $m \times m$ block as $b_i$. Here, we consider five options.

\textit{Downgrading} entails taking a weighted average or interpolation of neighboring pixels. This technique is already widely used: raw images are typically down- or up-sampled (as well as reshaped or cropped) to a standard size prior to applying the CNN. We do not consider DG as a form of LC since downgraded images are $l \times w \times c$ matrices of uncompressed pixels just like the uncompressed images. Rather, in this work, we use DG (as implemented in OpenCV's INTER\_AREA algorithm \cite{InterArea}) as the baseline against which LC is compared.

\textit{Principal Component Analysis} \cite{PCA} is a widely-used compression procedure in which an image is approximated as a linear combination of its $P$ principal eigenvectors. We apply PCA to $b_i$ and store the resulting parameters in an $n \times n$ matrix, padding with zeros as needed. We must therefore choose $P$ such that the number of PCA parameters does not exceed $n^2$. Concretely, we must require that:
\begin{equation} n \geq \sqrt{2mP + m} \label{eqn:PCA} \end{equation}
Though PCA has been widely studied, it offers two disadvantages. First, its computational complexity is very high, as eigenvectors must be calculated for each block. Second, Equation \ref{eqn:PCA} implies that PCA is simply incompatible with some dimensionalities. For example, it is impossible to represent even a single principal component of any $m \times m$ block in an $n \times n$ block if $m > 4$ and $n \leq 4$. For these reasons, we do not consider PCA further in this work.

\textit{Percentiles}. With percentile-based sampling, we sort the $m^2$ uncompressed points by their intensity and then sample from this distribution at pre-determined percentile values (e.g., the minimum, 33rd percentile, 67th percentile, and maximum). The computational complexity of this compression technique is somewhat high, as each block's $m^2$ values must be sorted. 

\textit{Random Matrix Multiplication (RMM)}. Some recent work in compressive sensing \cite{Thak} has looked at performing dimensionality reduction as a precursor to classification by multiplying the original features on the left by a (sparse) random matrix. In our case, we define $\vec{b}_i$ as the vectorized form of $b_i$. We then fill an $n^2 \times m^2$ matrix, $\mathcal{M}$, with values randomly drawn according to:
\begin{equation} \mathcal{M}_{ij} \sim \left\{ \begin{array}{cl} 0 & \textrm{with probability } 1-\gamma \\ \mathcal{N}(0, 1/{|\mathcal{M}|}) & \textrm{with probability } \gamma \\ \end{array} \right. \label{eqn:SRM} \end{equation}
where $0 \leq \gamma \leq 1$ (lower values of $\gamma$ are more efficient; in this work, we set $\gamma = 1$). We then perform dimensionality reduction according to $\vec{b}_i \rightarrow \mathcal{M} \vec{b}_i$. We then reshape the resulting vector to $n \times n$. 

\textit{Random Matrix Sketching (MS)}. Similar to RMM, MS fills an $n \times m$ matrix according to equation \eqref{eqn:SRM} (we again set $\gamma = 1$) and then compresses $b_i$ according to $b_i\rightarrow \mathcal{M} b_i \mathcal{M}^T$. This smaller dimensionality further reduces the compression technique's computational complexity. 

\section{Method}
\label{sec:Method}

Localized Compression entails using the techniques in Section \ref{sec:Theory} to compress entire $l \times w \times c$ images.  We refer to the entire uncompressed image as $x_i$. We begin by defining $m \times m$ blocks over $x_i$ (with $m \ll l, w$). We compress each channel separately and so consider only two dimensions here. In principal, these $m \times m$ blocks can be offset from one another by any number of pixels (i.e., an uncompressed pixel can be in zero, one, or multiple $m \times m$ blocks), but for simplicity, we consider only blocks that completely tile the image with no overlap (i.e., each uncompressed pixel is in exactly one $m \times m$ block). 

Algorithms 1 and 2 formally define LC for single-channel images (multi-channel images simply apply the compression operation to each channel separately). Algorithm 1 (``inline mode'') is a proof-of-concept in which the compression is performed at runtime: that is, we simply run the CNN as normal, inserting a step wherein we apply the compression operation to each $m \times m$ block and then apply the normal convolutional operation to the resulting $n \times n$ block. This is conceptually straightforward, but offers little or no savings in terms of storage efficiency (the uncompressed images must be stored) or computational efficiency (the reduction in learned convolutional parameters is roughly offset by the addition of compression operations). Algorithm 2 (``default mode'') makes some adjustments such that the compression is performed prior to runtime. Default mode achieves the same storage and computational efficiency as DG; however, this introduces some complications with respect to data augmentation (described below). 

Algorithm 1 (``inline mode'') begins by resizing ($\mathcal{R}$) each image to $\ell \times w$ and writing these images to disk. We then cycle through the images as normal. For each image, we use the data augmentation operations ($\mathcal{A}$) with randomly-drawn parameters to modify each image: these operations may include cropping to $M \times M$, taking a left-right flip, or any other data augmentation strategy. We then locally-compress each image ($\mathcal{C}$) and classify it using the CNN ($\mathcal{N}$). Note, when compressing with random matrices, we use the same random matrix for each image.

Algorithm 2 (``default mode'') differs in that it writes to disk \textit{after} performing the data augmentation and localized compression. It also requires that $s$ be divisible by $n$ so that the convolutional region will always stride over an integer number of compressed blocks. In this way, only the compressed images are written to disk (reducing the storage requirement), and the compression operations must only be performed once (reducing the computational requirement). The challenge with this ordering is that after compression, the full suite of data augmentation techniques can no longer be used; instead, only some limited set of data augmentation techniques ($\mathcal{A}_{lim}$) can be applied. In particular: 
\begin{itemize}
\item Crops. It is customary to take the final $M \times M$ crop during data augmentation (i.e., after resizing the image to $\ell \times w$). In default mode, this is still possible, however, the crops must not be allowed to sub-divide the $n \times n$ blocks. 
\item Flips. It is customary to take left-right flips of the image during data augmentation. In default mode, it is still possible to reverse the ordering of the $n \times n$ blocks; however, the internal structure of each block must not be changed. 
\end{itemize}
Other data augmentation schemes may or may not be applicable post-compression.

We therefore expect that networks trained in default mode will be somewhat less accurate than networks trained in inline mode. To bridge this gap, we allow default mode to produce $c$ copies of each image. These $c$ copies are produced using the full suite of data augmentation techniques ($\mathcal{A}$); at runtime, we randomly select one of these $c$ images and then apply the limited set of data augmentation techniques ($\mathcal{A}_{lim}$) to achieve further augmentation. We therefore expect that increasing $c$ will increase our classification accuracy, but will also increase our storage requirements. 

\begin{algorithm}
\caption{Classification with LC (inline mode)}
\begin{algorithmic}[1]
\renewcommand{\algorithmicrequire}{\textbf{Input:}}
\renewcommand{\algorithmicensure}{\textbf{Output:}}
\REQUIRE Image set $X$, network architecture $\mathcal{N}$
\ENSURE Label set $L={\ell_1, \ldots, \ell_Z}$
\HP $l$, $w$, $m$, $n$, $nEpochs$, $M$
 \FOR {$x_i \in X$}
   \STATE $x_i \leftarrow \mathcal{R}(l \times w) x_i$.
 \ENDFOR
 \STATE \textbf{Write} $X$.
 \FOR {$e \in$ $nEpochs$}
 \FOR {$x_i \in X$}
 \STATE {$x_i \leftarrow \mathcal{A}(M \times M) x_i$}
 \FOR {$i \in (0, l/m)$}
 \FOR {$j \in (0, w/m)$}
 \STATE {$x_c[i\cdot n:(i+1)\cdot n, j\cdot n:(j+1)\cdot n] \leftarrow \mathcal{C} x_i[i\cdot m:(i+1)\cdot m,j\cdot m:(j+1)\cdot m]$}
 \ENDFOR
 \ENDFOR
 \STATE {$\ell \leftarrow \mathcal{N} x_c$}
 \ENDFOR
 \ENDFOR
\RETURN $L$
\end{algorithmic}
\label{alg:LC_inline}
\end{algorithm}

\begin{algorithm}
\caption{Classification with LC (default mode)}
\begin{algorithmic}[1]
\renewcommand{\algorithmicrequire}{\textbf{Input:}}
\renewcommand{\algorithmicensure}{\textbf{Output:}}
\REQUIRE Image set $X$, network architecture $\mathcal{N}$
\ENSURE Label set $L={\ell_1, \ldots, \ell_Z}$
\HP $l$, $w$, $m$, $n$, $nEpochs$, $c$, $M$
 \FOR {$x_i \in X$}
 \FOR {$c_i \in c$}
   \STATE $x_i \leftarrow \mathcal{R}(l \times w) x_i$.
   \STATE $x_i \leftarrow \mathcal{A}(M \times M) x_i$.
   \FOR {$i \in (0, l/m)$}
   \FOR {$j \in (0, w/m)$}
     \STATE {$x_c[i\cdot n:(i+1)\cdot n, j\cdot n:(j+1)\cdot n] \leftarrow \mathcal{C} x_i[i\cdot m:(i+1)\cdot m,j\cdot m:(j+1)\cdot m]$}
   \ENDFOR
   \ENDFOR
 \ENDFOR
 \ENDFOR
 \STATE \textbf{Write} $X$.
 \FOR {$e \in$ $nEpochs$}
 \FOR {$x_i \in X$}
 \STATE {$x_i \leftarrow \mathcal{A}_{lim}(M \times M) x_i$}
 \STATE {$\ell \leftarrow \mathcal{N} x_i$}
 \ENDFOR
 \ENDFOR
\RETURN $L$
\end{algorithmic}
\label{alg:LC_default}
\end{algorithm}

\section{Numerical Experiments}
\label{sec:Experiments}
We test our procedure on two network architectures and two datasets. Our architectures are AlexNet \cite{AlexNet} and DenseNet \cite{DenseNet}: AlexNet is a dated architecture that has been widely used to evaluate compression algorithms, while DenseNet is a more modern architecture that achieves considerably higher accuracy. Both architectures require input images of a uniform size; we take $224 \times 224$ as the reference size for all images. Our datasets are the German Traffic Signs Recognition Benchmark (GTSRB) \cite{GTSRB} (39K training images over 37 classes) and ImageNet \cite{ImageNet} 2012 (1.3M training images over 1000 classes). While these are both standard datasets for classification challenges, a key difference is that most ImageNet images are larger than the reference size, whereas most GTSRB images are smaller than the reference size. We expect that LC will be more effective on large images (as there is more information to exploit).

We base our implementation of the networks, including parameters such as weight decay, on those from TensorFlow Slim \cite{TFSlim}, \cite{TF-DN}. In all tests (except where indicated), we begin by resizing and reshaping the images to $256 \times 256$ (e.g., $l=w=256$), cropping a random $224 \times 224$ patch from this (i.e., $M=224$), and then performing a left-right flip at random. All evaluation is performed with a single center crop and no left-right flip. We train all networks with the Momentum Optimizer with momentum 0.9 and a learning rate that begins at 0.01 and is reduced by an order of magnitude every 20 epochs, for a total of 65 epochs. This simple scheme is fully network-agnostic and offers relatively fast training times while giving top-1 accuracies only slightly lower than those reported by the network authors.

Our first test compare the percentile, RMM, and MS compression algorithms (as described in Section \ref{sec:Theory}) against the baseline of simply downgrading the images to the equivalent size. We use inline mode to allow all algorithms to use identical, off-the-shelf dataset augmentation techniques (as described in Section \ref{sec:Method}). We set $m=7$ and $n=2$; thus, the final images are $64 \times 64$. Note, these small images sizes require removing the last pooling layer from AlexNet. Our results, shown in Table \ref{tab:test1}, illustrate that LC is viable: all methods give results within a few percent of the baseline (DG), and the percentile method gives an accuracy 1-2\% higher than DG. We therefore perform LC with the percentile-based compression technique in the remainder of this work.

\begin{table*}
\centering
\caption{Test 1 Results: Localized Compression (inline mode) accuracy vs. compression algorithm}
\begin{tabular}{c|c|c|c|c|c}
\cline{3-6}
\multicolumn{2}{c}{} &  \multicolumn{2}{|c|}{ImageNet} & \multicolumn{2}{|c}{GTSRB} \\
\hline
Method & Size & AlexNet & DenseNet & AlexNet & DenseNet \\
\hline
Uncompressed & $224 \times 224$ & 56.8\% & 68.7\% & 96.8\% & 94.9\%\\
\hline
\hline
Downgraded & $64 \times 64$ & 27.0\% & 47.3\% & 92.8\% & 92.9\% \\
\hline
Percentiles & $64 \times 64$ & 29.4\% & 47.9\% & 94.3\% & 93.8\% \\
\hline
RMM & $64 \times 64$ & 26.4\% & 45.7\%  & 93.1\% & 94.0\% \\
\hline
MS & $64 \times 64$ & 26.4\% & 42.0\% & 93.4\% & 93.4\% \\
\hline
\end{tabular}
\label{tab:test1}
\end{table*}

Our second test validates default mode. In particular, we compare default mode's accuracy with various of $c$ against the accuracy achieved by inline mode. In addition to classification accuracy, Table \ref{tab:test2} also shows the uncompressed-to-compressed storage ratio (SR) and the uncompressed-to-compressed computational ratio (CR). The results show that LC is still viable in default mode: even with $c=1$, LC remains more accurate than DG. Setting $c=2$ further increases the classification accuracy; however, continuing to increase $c$ shows only a modest improvement in classification accuracy. In the remainder of this work, we use default mode with $c=2$. 

Third, we test LC for different compression ratios. In particular, we keep $n=2$ and vary the size of $m$; larger values of $m$ therefore result in smaller compressed image sizes. Our results for LC and DG are given in Table \ref{tab:test3}, and show that LC consistently outperforms DG for significant compression ratios (i.e., reducing the number of pixels by more than a factor of 4), but advantage is less clear for smaller compression ratios. 
\begin{table*}
\centering
\caption{Test 2 Results: Localized Compression (default mode) accuracy vs. $c$}
\begin{tabular}{c|c|c|c|c|c|c}
\cline{4-7}
\multicolumn{3}{c}{} &  \multicolumn{2}{|c|}{ImageNet} & \multicolumn{2}{|c}{GTSRB} \\
\hline
$c$ & CR & SR & AlexNet & DenseNet & AlexNet & DenseNet \\
\hline
1 & 12.25x & 12.25x & 28.4\% & 47.4\% & 94.5\% & 94.4\% \\
\hline
2 & 12.25x & 6.125x & 28.9\% & 47.9\% & 94.1\% & 93.5\% \\
\hline
4 & 12.25x & 3.06x & 29.0\% & 48.0\% & 94.4\% & 93.7\% \\
\hline
\hline
inline & 1x  & 1x  & 29.4\% & 47.9\% & 94.3\% & 93.8\% \\
\hline
\end{tabular}
\label{tab:test2}
\end{table*}

\begin{table*}
\centering
\caption{Test 3 Results: Localized Compression (default mode) accuracy vs. compression ratio}
\begin{tabular}{c|c|c|c|c|c|c|c|c}
\cline{2-9}
&  \multicolumn{4}{|c|}{ImageNet} & \multicolumn{4}{|c}{GTSRB} \\
\cline{2-9}
& \multicolumn{2}{|c|}{AlexNet} & \multicolumn{2}{|c|}{DenseNet} & \multicolumn{2}{|c|}{AlexNet} & \multicolumn{2}{|c}{DenseNet}\\
\cline{1-9}
Compression Ratio & LC & DG & LC & DG & LC & DG & LC & DG \\
\hline
$8 \times 8 \rightarrow 2 \times 2$  &  27.6\% & 25.6\%  & 42.7\% & 42.3\% & 94.0\% & 93.2\% & 93.7\% & 93.5\% \\
\hline
$7 \times 7 \rightarrow 2 \times 2$  & 28.9\%  & 27.0\%  & 47.9\% & 47.3\% & 94.1\% & 92.8\% & 93.5\% & 92.9\% \\
\hline
$6 \times 6 \rightarrow 2 \times 2$  & 31.8\% & 30.7\% & 50.2\% & 50.2\% & 94.0\% & 94.1\% & 94.0\% & 93.3\% \\
\hline
$5 \times 5 \rightarrow 2 \times 2$  & 37.0\%  & 36.1\% & 53.0\% & 53.2\% & 94.9\% & 94.4\% & 94.7\% & 93.7\% \\
\hline
$4 \times 4 \rightarrow 2 \times 2$  & 44.2\% & 45.4\%  & 58.1\% & 61.2\% & 95.5\% & 95.0\% & 94.3\% & 94.2\% \\
\hline
$3 \times 3 \rightarrow 2 \times 2$  & 52.5\% & 52.2\% & 62.6\% & 63.4\% & 96.0\% & 97.0\% & 94.3\% & 94.0\% \\
\hline
\end{tabular}
\label{tab:test3}
\end{table*}

Finally, we consider applying LC to larger images: rather than beginning with $224 \times 224$ and compressing, we begin with large $784 \times 784$ images and compress to $224 \times 224$. To evaluate this, we consider only the ImageNet dataset, and select only those images with more than $784^2$ pixels (of which there are 30,192 for training and 1,110 for testing). Our results are given in Table \ref{tab:test4}. Though the CNNs are clearly data starved, our results suggest that LC gives higher accuracy than DG for larger images just as it did for smaller images. 

\begin{table}
\centering
\caption{Test 4 Results: Localized Compression vs. Downsampling on $784 \times 784$ ImageNet images}
\begin{tabular}{c|c|c}
\cline{2-3}
& AlexNet & DenseNet \\
\hline
DG & 16.7\% & 15.5\% \\
\hline
LC & 17.2\% & 17.8\% \\
\hline
\end{tabular}
\label{tab:test4}
\end{table}

\section{Applying random matrices to fully-connected layers}
\label{sec:Digression}

We now take a brief digression to consider a related problem: using the techniques of Section \ref{sec:Theory} to compress the fully-connected layers inside the CNN itself. As discussed in Section \ref{sec:Related Work}, substantial work has been put into compressing these fully-connected layers in deep neural networks; our contribution is to extend this work by applying static, non-circulant random matrices to CNNs. Our strategy is to introduce a new deterministic layer immediately prior to each fully-connected layer that compresses the inputs to the following layer. The percentile-based sampling method, though effective for LC, is not an appropriate choice for this layer, as it would continually reorder the features. We therefore select the MS-based technique for this layer; this efficiently and dramatically reduces the number of weights in the hidden layer. 

As an numerical example, we again consider the AlexNet architecture, which contains three fully-connected layers, the first of which contains 6400 weights. We then reshape these weights to $25 \times 256$ and multiply on the left by a $13 \times 25$ random matrix. This produces 3328 weights, which we feed into the next fully-connected layer. We repeat this for each fully-connected layers. We can also reduce the number of nodes in each FC layer to compensate for the reduced number of inputs.

\begin{table}
\centering
\caption{Results: MS compression of each FC layer in AlexNet}
\begin{tabular}{c|c|c|c|c|c|c}
\hline
\multicolumn{3}{c|}{CR by Layer} & &  & & \\
\cline{1-3}
FC1 & FC2 & FC3 & nNodes & GTSRB & ImageNet & CR \\
\hline
1.00  & 1.00 & 1.00 & 4096   & 96.8\% & 58.4\% & 1.00 \\
\hline
0.50  & 1.00 & 1.00 & 4096   & 96.5\% & 57.5\% & 0.59 \\
\hline
0.50  & 0.50 & 0.50 & 4096  & 96.0\% &  57.8\% & 0.55 \\
\hline
0.50 & 0.50 & 0.50 & 2048   &  98.2\% & 54.3\% & 0.28  \\
\hline
\end{tabular}
\label{tab:FC}
\end{table}

Table \ref{tab:FC} shows our results applying this to the GTSRB and ImageNet datasets with the AlexNet network architecture (we did not consider DenseNet, as it contains only a single fully-connected layer). Table \ref{tab:FC} reports the compression ratio for each of the three fully-connected layers (FC1, FC2, and FC3), and the number of nodes contained in each FC layer (nNodes). Compared to the uncompressed AlexNet, we observe a 1\% increase in accuracy on GSTRB when compressing the network up to a 72\% and less than  1\% decrease in accuracy when tested using ImageNet compressing the network by 45\%. The counter-intuitive increase in accuracy for GTSRB may be because the original images are so small that the large network has too many parameters relative to the images' information content.

\section{Conclusions}
\label{sec:Conclusions}
The primary contribution of this work is to introduce Localized Compression (LC), an alternative to downgrading when CNNs require an image size much smaller than the original image's resolution. In some sense, LC is a generalization of downgrading: downgrading always performs some sort of pixel averaging and requires $n=1$, whereas LC supports different compression techniques and different values of $n$. The most successful compression technique, percentile-based sampling, could be viewed as applying a generalization of a pooling layer to the original image. By choosing $n$ such that $n$ divides $r$, LC supports any network architecture.  

We also extended previous work \cite{Thak}, \cite{RandProjPriorWork} on applying sparse random matrices to deep neural networks. Though percentile-based sampling outperformed random-matrix-based techniques on LC, we showed that sparse random matrices are an effective way to compress both convolutional and fully-connected layers in CNNs.

With respect to LC, our results show that when it is used with percentile-based sampling and relatively high compression ratios, LC gives a 1-2\% accuracy improvement over downgrading. LC can therefore be useful in applications where the average image size is much larger than a CNN's reference size. This is potentially a useful capability: many modern cameras can produce high-resolution images; LC provides a way to exploit the extra information that such devices provide without increasing the computational or SWaP requirements. 

\section*{Acknowledgment}

The authors would like to acknowledge Profs. Pramod Varshney and Thakshila Wimalajeewa at Syracuse University for sharing their expertise with using random matrices for classification.

This work was supported by  the Air Force Research Laboratory under contract FA8650-17-C-1154. 

\bibliography{paper}
\bibliographystyle{IEEEtran}

\end{document}